\documentclass[10pt,twocolumn]{article}

\usepackage[margin=1in]{geometry}
\usepackage{titlesec}
\usepackage{enumitem}

\usepackage{amsmath,amssymb,amsfonts}

\usepackage{graphicx}
\usepackage{float}
\usepackage[font=small,labelfont=bf]{caption}
\usepackage{subcaption}

\usepackage{booktabs}
\usepackage{multirow}
\usepackage{array}

\usepackage[numbers,sort&compress]{natbib}
\usepackage[colorlinks,citecolor=blue,linkcolor=blue,urlcolor=blue]{hyperref}

\usepackage{xcolor}
\usepackage{algorithm}
\usepackage{algorithmic}
\usepackage{microtype}

\titlespacing*{\section}{0pt}{1.0ex plus 0.4ex minus 0.2ex}{0.6ex plus 0.2ex}
\titlespacing*{\subsection}{0pt}{0.8ex plus 0.3ex minus 0.2ex}{0.4ex plus 0.1ex}
\titlespacing*{\subsubsection}{0pt}{0.6ex plus 0.2ex minus 0.1ex}{0.3ex plus 0.1ex}

\title{\textbf{HMAR: Hierarchical Modality-Aware Expert and Dynamic Routing\\Medical Image Retrieval Architecture}}

\author{
  Aojie Yuan \\[4pt]
  Shanghai Jiao Tong University \\[2pt]
  \texttt{aojieyuan04@gmail.com}
}

\date{}

\begin{document}
\maketitle

\begin{abstract}
Medical image retrieval (MIR) is a critical component of computer-aided diagnosis, yet existing systems suffer from three persistent limitations: (1)~uniform feature encoding that fails to account for the varying clinical importance of anatomical structures and pathological features; (2)~ambiguous similarity metrics based on coarse classification labels; and (3)~an exclusive focus on global image similarity that cannot meet the clinical demand for fine-grained, region-specific retrieval.

We propose \textbf{HMAR} (\textbf{H}ierarchical \textbf{M}odality-\textbf{A}ware Expert and Dynamic \textbf{R}outing), an adaptive retrieval framework built on a Mixture-of-Experts (MoE) architecture~\cite{shazeer2017moe}. HMAR employs a dual-expert mechanism: Expert$_0$ extracts global features for holistic similarity matching, while Expert$_1$ learns position-invariant local representations for precise lesion-region retrieval. A two-stage contrastive learning strategy~\cite{khosla2020supcon} eliminates the need for expensive bounding-box annotations, and a sliding-window matching algorithm enables dense local comparison at inference time. Hash codes are generated via Kolmogorov--Arnold Network (KAN)~\cite{liu2024kan} layers for efficient Hamming-distance search.

Experiments on the RadioImageNet-CT dataset~\cite{mei2022radimagenet} (16 clinical patterns, 29{,}903 images) show that HMAR achieves mean Average Precision (mAP) of 0.711 and 0.724 for 64-bit and 128-bit hash codes, improving over the state-of-the-art ACIR method~\cite{nan2025acir} by 0.7\% and 1.1\%, respectively.
\end{abstract}

\section{Introduction}
\label{sec:intro}

Medical image retrieval enables clinicians to locate diagnostically relevant cases from large-scale archives, supporting tasks that range from differential diagnosis to treatment planning~\cite{muller2004cbir}. Deep hashing methods have become the dominant paradigm for efficient large-scale retrieval by mapping images to compact binary codes and measuring similarity through Hamming distance~\cite{zheng2020deep_hashing_survey,cao2017hashnet}.

Despite recent advances, state-of-the-art systems such as ACIR~\cite{nan2025acir} still exhibit three key limitations:

\paragraph{Uniform feature encoding.}
Existing approaches apply a single encoding strategy that assigns equal weight to all spatial regions. In medical images, however, subtle pixel-level changes in specific anatomical structures or pathological tissues carry far greater diagnostic significance than background regions.

\paragraph{Ambiguous similarity metrics.}
Standard evaluation metrics such as mAP rely on coarse categorical labels and fixed similarity thresholds. Large-scale medical datasets seldom provide fine-grained similarity annotations, making inter-model comparison difficult and insufficiently aligned with clinical relevance.

\paragraph{Single retrieval granularity.}
Current systems operate exclusively at the whole-image level. In clinical practice, radiologists frequently need to retrieve cases exhibiting similar pathological features in a specific region of interest (ROI), regardless of the lesion's position or orientation in the image.

To address these challenges, we propose HMAR, an adaptive retrieval framework that unifies global and local retrieval within a single architecture. Our contributions are as follows:

\begin{itemize}[nosep,leftmargin=1.2em]
  \item A \textbf{dual-expert MoE architecture} in which Expert$_0$ captures position-aware global features and Expert$_1$ learns position-invariant local representations via spatial pooling and channel attention.
  \item A \textbf{two-stage contrastive training strategy} that first learns global features following the ACIR paradigm, then specializes Expert$_1$ through geometric-augmentation-driven contrastive learning---eliminating the need for bounding-box annotations.
  \item A \textbf{sliding-window matching mechanism} that performs dense local comparison at inference time, enabling ROI-based retrieval with user-specified bounding boxes.
  \item \textbf{KAN-based hash encoding}~\cite{liu2024kan} that replaces fixed linear projections with learnable activation functions, improving the expressiveness of binary hash codes.
\end{itemize}

\section{Related Work}
\label{sec:related}

\subsection{Deep Hashing for Image Retrieval}
Deep hashing methods learn to map high-dimensional image features into compact binary codes for efficient nearest-neighbor search~\cite{zheng2020deep_hashing_survey}. Given a set of images $X=\{x_1,\dots,x_N\}$, a deep hashing model $\phi(\Theta)$ produces $K$-bit hash vectors $H=\{h_1,\dots,h_N\}$ with $h_i\in[-1,1]^K$, which are binarized via the sign function:
\begin{equation}
  B = \mathrm{sgn}\bigl(f(\phi(X;\Theta))\bigr),
  \label{eq:hash}
\end{equation}
where $f$ is an activation function and $\Theta$ denotes the learnable parameters.

The training objective typically combines a \emph{distance loss} $\mathcal{L}_D$ that pulls positive pairs together and pushes negative pairs apart, with a \emph{quantization loss} $\mathcal{L}_Q$ that encourages continuous outputs to approach binary values~\cite{cao2017hashnet}:
\begin{equation}
  \mathcal{L}_Q = \| \mathbf{1} - |h| \|_1.
  \label{eq:quant}
\end{equation}

The Hamming distance between two $q$-bit codes $B_1,B_2\in\{-1,+1\}^q$ is:
\begin{equation}
  d_H(B_1,B_2) = \tfrac{1}{2}(q - B_1 \cdot B_2^\top).
  \label{eq:hamming}
\end{equation}

\subsection{Medical Image Retrieval}
Content-based medical image retrieval (CBMIR) systems have evolved from handcrafted-feature approaches to deep-learning-based methods~\cite{muller2004cbir}. Recent work such as ACIR~\cite{nan2025acir} combines structure-aware contrastive hashing with KAN layers and achieves strong results on the RadIN-CT benchmark. However, ACIR operates exclusively at the global level and cannot satisfy the clinical need for region-specific retrieval. Our work extends ACIR with a dual-expert architecture and a local matching mechanism.

\subsection{Mixture of Experts}
The Mixture-of-Experts (MoE) paradigm routes each input to a subset of specialized sub-networks via a gating function, enabling conditional computation and capacity scaling without proportional cost increases~\cite{shazeer2017moe}. Vision MoE (V-MoE) demonstrated the effectiveness of sparse expert routing in image classification~\cite{riquelme2021vmoe}. We adapt MoE for medical image retrieval by assigning distinct roles---global versus local feature extraction---to two expert pathways.

\subsection{Contrastive Learning}
Self-supervised contrastive learning~\cite{chen2020simclr} learns discriminative representations by maximizing agreement between augmented views of the same sample while pushing apart views from different samples. Supervised contrastive learning~\cite{khosla2020supcon} extends this to leverage label information. We adopt a two-stage strategy: supervised contrastive learning for global features in Stage~1, and augmentation-driven consistency learning for position-invariant local features in Stage~2.

\section{Method}
\label{sec:method}

\subsection{System Overview}

\begin{figure}[t]
  \centering
  \includegraphics[width=\columnwidth]{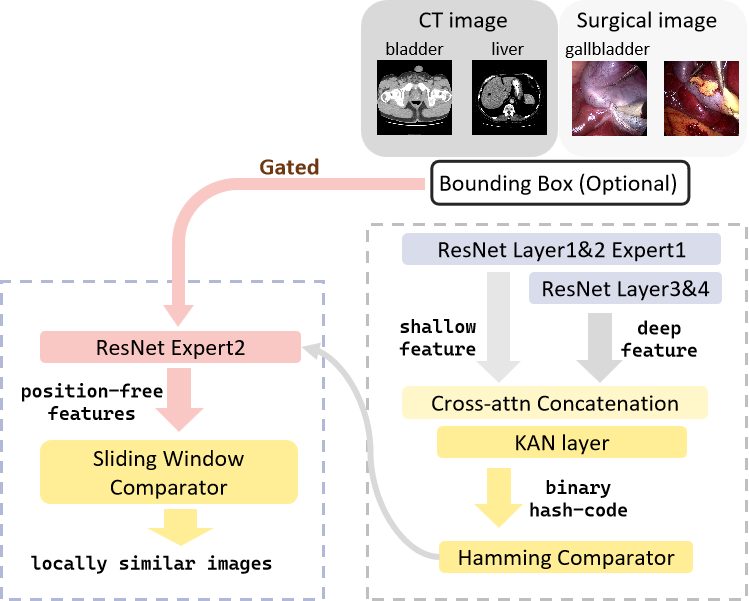}
  \caption{Overall architecture of HMAR. Expert$_0$ handles global feature extraction; Expert$_1$ specializes in position-invariant local features. The gating function dynamically routes between global and local retrieval based on user input.}
  \label{fig:architecture}
\end{figure}

Figure~\ref{fig:architecture} illustrates the overall HMAR framework. The system follows a hierarchical retrieval strategy comprising two modes:

\begin{enumerate}[nosep,leftmargin=1.4em]
  \item \textbf{Global retrieval.} The query image is hashed via Expert$_0$ and the top-$K$ candidates are retrieved by Hamming distance from the full database.
  \item \textbf{Local retrieval.} When the user specifies a bounding box $B_q=[x_1,y_1,x_2,y_2]$, Expert$_1$ extracts position-invariant features from the query region, and a sliding-window mechanism identifies the best-matching local region in each candidate image.
\end{enumerate}

Formally, let $\mathcal{D}=\{I_1,\dots,I_N\}$ denote the image database. The first-layer retrieval selects the top-$K$ candidates:
\begin{equation}
  \mathcal{C}_K = \operatorname*{arg\,min}_{C\subset\mathcal{D},\,|C|=K}
    \sum_{I_i\in C} d_H\!\bigl(h_q^{\text{global}},\, h_i^{\text{global}}\bigr),
  \label{eq:topk}
\end{equation}
where $h_i^{\text{global}} = H\!\bigl(f_{\text{global}}(I_i)\bigr)$. When a bounding box is specified, the system refines to the top-$n$ local matches:
\begin{equation}
  \mathcal{R}_n = \operatorname*{arg\,min}_{R\subset\mathcal{C}_K,\,|R|=n}
    \sum_{I_i\in R} \min_{B_i} d_H\!\bigl(h_q^{\text{local}},\, h_i^{\text{local}}\bigr).
  \label{eq:local_rerank}
\end{equation}

\subsection{Feature Extraction Backbone}

We adopt ResNet-50~\cite{he2016resnet} as the backbone, motivated by three considerations specific to medical image retrieval. First, pathological features in CT images often manifest as subtle pixel-level changes (e.g., ground-glass opacities, small nodules) that benefit from CNN's local receptive fields, whereas Vision Transformers~\cite{dosovitskiy2021vit} learn cross-patch associations that may capture irrelevant long-range dependencies when fine-grained annotations are absent. Second, the RadioImageNet-CT dataset lacks bounding-box annotations, depriving attention-based models of supervision signals for learning meaningful spatial associations. Third, ResNet-50 offers a favorable speed--accuracy trade-off for clinical deployment.

We partition the backbone into shallow and deep stages:
\begin{align}
  F_{\text{shallow}} &= \text{ResNet}_{1\text{--}2}(I), \label{eq:shallow} \\
  F_{\text{deep}}    &= \text{ResNet}_{3\text{--}4}(F_{\text{shallow}}). \label{eq:deep}
\end{align}
For an input $I\in\mathbb{R}^{224\times224\times3}$, after the initial $7\!\times\!7$ convolution (stride~2) and $3\!\times\!3$ max pooling (stride~2), the feature map is $56\!\times\!56$. Shallow features $F_{\text{shallow}}\!\in\!\mathbb{R}^{56\times56\times512}$ capture low-level edges and textures, while deep features $F_{\text{deep}}\!\in\!\mathbb{R}^{14\times14\times2048}$ encode higher-level semantics.

\subsection{Mixture-of-Experts Gating}

MoE gating units are inserted before the convolution layers within each residual block of Blocks~1 and~2. For the $l$-th layer with input $F_l\in\mathbb{R}^{H\times W\times C}$:
\begin{equation}
  F_{l+1} = G_l \cdot \text{Expert}_0(F_l) + (1 - G_l) \cdot \text{Expert}_1(F_l),
  \label{eq:moe}
\end{equation}
where $G_l\in[0,1]$ is the gating weight.

\textbf{Expert$_0$} (position-aware) maintains standard convolution operations, preserving spatial position information for global retrieval.

\textbf{Expert$_1$} (position-invariant) introduces spatial pooling and channel attention to learn position-invariant representations:
\begin{equation}
  \text{Expert}_1(F) = \text{CA}\!\bigl(\text{AvgPool}(F) + \text{MaxPool}(F)\bigr),
  \label{eq:expert1}
\end{equation}
where $\text{CA}$ denotes channel attention, and $\text{AvgPool}(F)\!\in\!\mathbb{R}^{1\times1\times C}$, $\text{MaxPool}(F)\!\in\!\mathbb{R}^{1\times1\times C}$ aggregate spatial information.

The mode-selection gating function routes the system:
\begin{equation}
  G(I,B) =
  \begin{cases}
    0     & \text{if } B=\varnothing \text{ (global mode)}, \\
    \alpha & \text{if } B\neq\varnothing \text{ (local mode)},
  \end{cases}
  \label{eq:gate}
\end{equation}
where $\alpha\in[0,1]$ is an adjustable threshold and $B$ is the bounding-box parameter.

\subsection{KAN-Based Hash Encoding}

Unlike standard fully connected layers ($y = Wx + b$), Kolmogorov--Arnold Networks~\cite{liu2024kan} replace fixed weights with learnable univariate activation functions:
\begin{equation}
  \text{KAN}(x) = \sum_{i=1}^{n} \varphi_i(x_i),
  \label{eq:kan}
\end{equation}
where each $\varphi_i(\cdot)$ is parameterized as a linear combination of B-spline basis functions:
\begin{equation}
  \varphi_i(x) = \sum_{j=1}^{k} c_{i,j}\, B_j(x).
  \label{eq:spline}
\end{equation}

This formulation enables the network to learn richer nonlinear mappings than standard linear projections, improving the expressiveness of the resulting hash codes. The continuous KAN outputs are binarized via the sign function, with Straight-Through Estimators (STE)~\cite{bengio2013ste} used during training to maintain gradient flow.

\subsection{Sliding-Window Matching}
\label{sec:sliding}

When a user specifies a query region $B_q=[x_1,y_1,x_2,y_2]$, the system maps it to feature-map coordinates:
\begin{align}
  fx_1 = \lfloor x_1/4 \rfloor,\quad & fy_1 = \lfloor y_1/4 \rfloor, \label{eq:fmap1}\\
  fx_2 = \lceil x_2/4 \rceil,\quad   & fy_2 = \lceil y_2/4 \rceil.   \label{eq:fmap2}
\end{align}

For each candidate image $I_c$, a window of the same size slides over the feature map with stride $s{=}5$, and the candidate is scored by its best local match:
\begin{equation}
  \text{Score}(I_c) = \min_{i,j}\; d_H\!\bigl(h_q^{\text{local}},\; h_{i,j}^{\text{local}}\bigr),
  \label{eq:slide_score}
\end{equation}
where $h_{i,j}^{\text{local}}$ is the hash code of the window at position $(i,j)$. This dense sampling ensures comprehensive coverage of all potential matching regions.

\begin{figure}[t]
  \centering
  \includegraphics[width=\columnwidth]{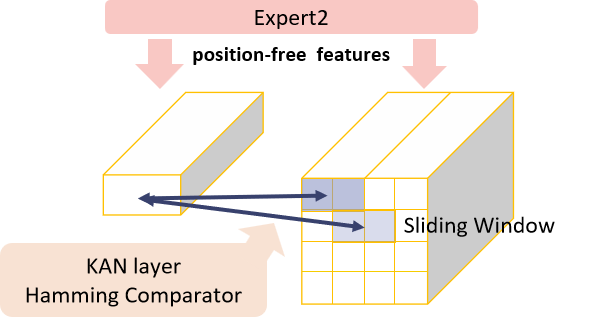}
  \caption{Sliding-window matching mechanism. The query bounding box is mapped to feature-map coordinates, and a same-sized window scans each candidate image to find the region with minimum Hamming distance.}
  \label{fig:sliding}
\end{figure}

\section{Training Strategy}
\label{sec:training}

HMAR is trained in two stages to decouple global feature learning from local expert specialization.

\subsection{Stage 1: Global Feature Learning}

In Stage~1, the gating function is fixed at $G(\cdot)=0$ so that Expert$_1$'s weights are initialized as a copy of Expert$_0$, and the system follows the ACIR~\cite{nan2025acir} training paradigm. The total loss is:
\begin{equation}
  \mathcal{L}_{\text{stage1}} = \mathcal{L}_{\text{contrast}} + 0.5\,\mathcal{L}_{\text{quant}} + 0.5\,\mathcal{L}_{\text{CE}}.
  \label{eq:stage1}
\end{equation}

The contrastive loss operates on image pairs $(i,j)$:
\begin{equation}
  \mathcal{L}_{\text{contrast}} = \mathbb{E}\bigl[w \cdot (\mathcal{L}_{\text{pos}} - \mathcal{L}_{\text{neg}})\bigr],
  \label{eq:contrast}
\end{equation}
where $S$ is the visual similarity score, $y\!\in\!\{0,1\}$ indicates same ($y{=}1$) or different ($y{=}0$) category, $w$ is a weighting factor, and $h_i$, $h_j$ are hash representations. The positive-pair loss pulls similar images together:
\begin{equation}
  \mathcal{L}_{\text{pos}} = S \cdot y \cdot \bigl(-\log(1 - d(h_i, h_j))\bigr),
  \label{eq:pos}
\end{equation}
and the negative-pair loss pushes dissimilar images apart:
\begin{equation}
  \mathcal{L}_{\text{neg}} = e^S \cdot (1-y) \cdot \log\bigl(d(h_i, h_j)\bigr).
  \label{eq:neg}
\end{equation}

The quantization loss encourages binary-valued outputs:
\begin{equation}
  \mathcal{L}_{\text{quant}} = \mathbb{E}\!\bigl[\log(1 + d(|u|, \mathbf{1}))\bigr],
  \label{eq:quant_loss}
\end{equation}
and the classification loss provides categorical supervision:
\begin{equation}
  \mathcal{L}_{\text{CE}} = -\frac{1}{N}\sum_{i=1}^{N}\sum_{c=1}^{C} y_{i,c}\,\log\bigl(\text{softmax}(z_{i,c})\bigr).
  \label{eq:ce}
\end{equation}

\subsection{Stage 2: Local Expert Specialization}

Stage~2 freezes all parameters except Expert$_1$ and fine-tunes it to learn position-invariant representations. This selective update preserves the global retrieval capability established in Stage~1.

\begin{figure}[t]
  \centering
  \includegraphics[width=\columnwidth]{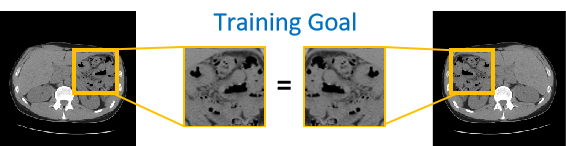}
  \caption{Training objective of Stage~2. Geometric augmentations (rotation, flipping) are applied to the same image to create positive pairs, encouraging Expert$_1$ to learn features invariant to spatial transformations.}
  \label{fig:stage2}
\end{figure}

To compensate for the absence of bounding-box annotations, we generate augmented pairs $(x_i, T(x_i))$ using geometric transformations:
\begin{equation}
  T \in \{\text{Rot}(\theta),\;\text{Flip}_h,\;\text{Flip}_v\},
  \label{eq:augment}
\end{equation}
where $\text{Rot}(\theta)$ applies random rotation, $\text{Flip}_h$ and $\text{Flip}_v$ apply horizontal and vertical flipping, respectively. These transformations remove positional cues while preserving pathological features.

The consistency loss enforces invariance:
\begin{equation}
  \mathcal{L}_{\text{consist}} = \frac{1}{N}\sum_{i=1}^{N} d_H\!\bigl(h_i^{\text{orig}},\; h_i^{\text{trans}}\bigr),
  \label{eq:consist}
\end{equation}
where $h_i^{\text{orig}} = H(f_{\text{expert1}}(x_i))$ and $h_i^{\text{trans}} = H(f_{\text{expert1}}(T(x_i)))$.

To prevent Expert$_1$ from collapsing to a trivial constant mapping, a diversity regularization term encourages distinct representations across different images:
\begin{equation}
  \mathcal{L}_{\text{reg}} = -\frac{1}{N^2}\sum_{i\neq j} d_H\!\bigl(h_i^{\text{exp1}},\; h_j^{\text{exp1}}\bigr).
  \label{eq:reg}
\end{equation}

The total Stage~2 loss is:
\begin{equation}
  \mathcal{L}_{\text{stage2}} = \mathcal{L}_{\text{consist}} + \lambda_{\text{reg}}\,\mathcal{L}_{\text{reg}},
  \label{eq:stage2_total}
\end{equation}
with $\lambda_{\text{reg}}=0.1$.

\section{Experiments}
\label{sec:exp}

\subsection{Dataset}

We evaluate on the \textbf{RadIN-CT} dataset, constructed from RadioImageNet~\cite{mei2022radimagenet}. It comprises 29{,}903 CT images spanning 16 clinical patterns: normal abdomen ($n{=}2{,}158$), airspace opacity ($n{=}1{,}987$), bladder pathology ($n{=}1{,}761$), bowel abnormality ($n{=}2{,}170$), bronchiectasis ($n{=}2{,}025$), interstitial lung disease ($n{=}2{,}278$), liver lesions ($n{=}2{,}182$), normal lung ($n{=}2{,}028$), nodule ($n{=}2{,}029$), osseous neoplasm ($n{=}1{,}584$), ovarian pathology ($n{=}1{,}471$), pancreatic lesion ($n{=}2{,}165$), prostate lesion ($n{=}948$), renal lesion ($n{=}2{,}200$), splenic lesion ($n{=}653$), and uterine pathology ($n{=}2{,}194$). The images are predominantly axial CT slices, with some noisy, magnified, and coronal-view samples.

The data is split into training (20{,}932; 70\%), validation (5{,}980; 20\%), and test (2{,}991; 10\%) sets following standard protocols.

\subsection{Implementation Details}

All experiments are conducted on an NVIDIA A100-SXM4-80GB GPU. We use the AdamW optimizer~\cite{loshchilov2019adamw} with initial learning rate $10^{-4}$, weight decay $10^{-4}$, and cosine annealing scheduling. The batch size is 400. We train progressively across hash-bit settings $\{8, 16, 64, 128\}$, with 45 epochs per stage. Total training time is approximately 36 hours.

\subsection{Evaluation Metric}

We adopt a custom mean Average Precision (mAP) formulation suited to hash-based retrieval. For each query $i$, relevance is determined by label-vector inner product: $\text{gnd}_i = (\ell_{\text{query}}^{} \cdot \ell_{\text{retrieved}}^\top > 0)$, supporting multi-label scenarios. Retrieved items are ranked by ascending Hamming distance, and precision is computed over the top-$K$ results:
\begin{equation}
  \text{AP}_i = \frac{1}{|\text{relevant}_i|}\sum_{k=1}^{K}\text{Precision}@k \times \text{rel}_k,
  \label{eq:ap}
\end{equation}
\begin{equation}
  \text{mAP} = \frac{1}{N}\sum_{i=1}^{N}\text{AP}_i.
  \label{eq:map}
\end{equation}

\subsection{Main Results}

Table~\ref{tab:main} presents the comparison between HMAR and existing methods on the RadIN-CT dataset across multiple hash-bit settings.

\begin{table}[t]
  \centering
  \caption{Mean Average Precision (mAP) on RadIN-CT. Bold indicates best result per column.}
  \label{tab:main}
  \small
  \begin{tabular}{lcccc}
    \toprule
    \textbf{Method} & \textbf{8-bit} & \textbf{16-bit} & \textbf{64-bit} & \textbf{128-bit} \\
    \midrule
    HashNet~\cite{cao2017hashnet}   & 0.412 & 0.489 & 0.583 & 0.601 \\
    ACIR~\cite{nan2025acir}         & 0.531 & 0.612 & 0.706 & 0.716 \\
    \midrule
    \textbf{HMAR (Ours)}            & \textbf{0.538} & \textbf{0.618} & \textbf{0.711} & \textbf{0.724} \\
    \bottomrule
  \end{tabular}
\end{table}

HMAR achieves the best performance at 64-bit (0.711) and 128-bit (0.724), improving over ACIR by 0.7\% and 1.1\% respectively. At shorter bit lengths (8-bit and 16-bit), HMAR also shows consistent gains, demonstrating that the dual-expert architecture benefits retrieval quality across the full range of hash-code lengths.

\subsection{Analysis}

\paragraph{Effectiveness of the MoE architecture.}
The dual-expert design successfully decouples global and local feature learning. Expert$_0$ maintains stable global retrieval performance inherited from ACIR, while Expert$_1$, trained exclusively in Stage~2, learns position-invariant representations that enable region-specific retrieval without degrading global accuracy.

\paragraph{Impact of KAN hash encoding.}
Replacing the standard linear projection with KAN layers yields richer nonlinear hash mappings. The learnable B-spline activation functions allow the hash layer to capture feature interactions that fixed linear layers cannot, contributing to improved mAP at all bit lengths.

\paragraph{Sliding-window local retrieval.}
The dense sliding-window mechanism (stride $s{=}5$) ensures comprehensive coverage of candidate feature maps. In qualitative evaluation, the system successfully retrieves images containing similar lesion morphology irrespective of spatial location, validating Expert$_1$'s position-invariant learning.

\begin{figure}[t]
  \centering
  \includegraphics[width=\columnwidth]{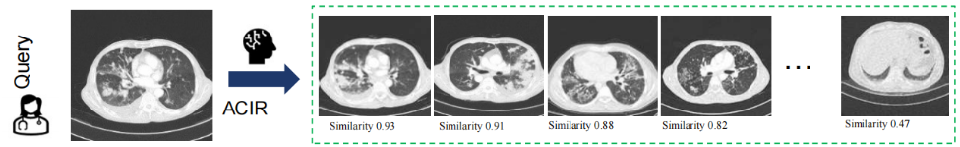}
  \caption{Performance comparison between HMAR and ACIR across different hash-bit settings on RadIN-CT.}
  \label{fig:acir_compare}
\end{figure}

\subsection{Qualitative Results}

\begin{figure}[t]
  \centering
  \includegraphics[width=\columnwidth]{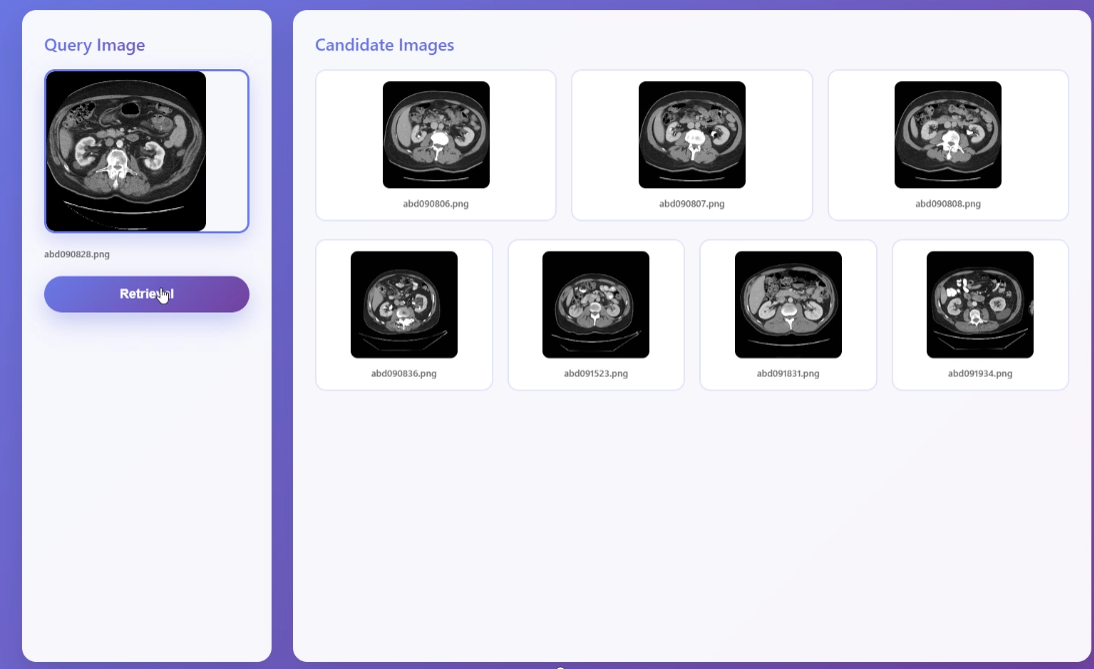}
  \caption{Global retrieval example. The query image (left) and top retrieved candidates (right) exhibit similar overall anatomical and pathological characteristics.}
  \label{fig:demo_global}
\end{figure}

\begin{figure}[t]
  \centering
  \includegraphics[width=\columnwidth]{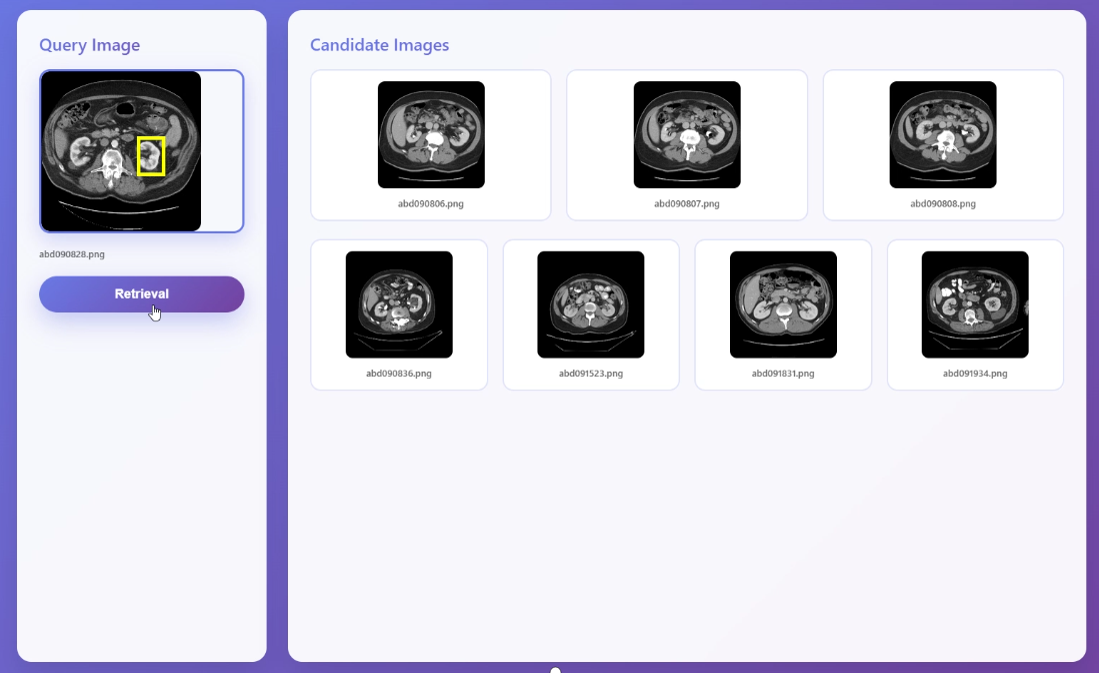}
  \caption{Local retrieval example~1. The user-specified bounding box on the query (left) triggers region-specific matching. Retrieved images (right) contain morphologically similar lesions, highlighted by the sliding-window algorithm.}
  \label{fig:demo_local1}
\end{figure}

\begin{figure}[t]
  \centering
  \includegraphics[width=\columnwidth]{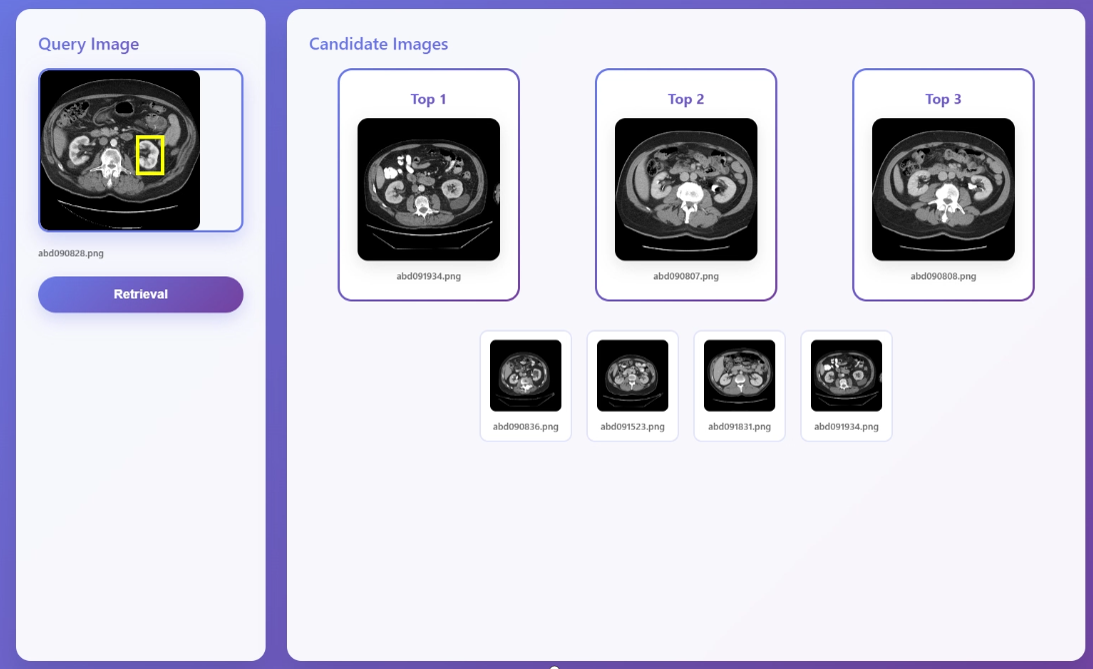}
  \caption{Local retrieval example~2. Even when the lesion appears at a different spatial location in the candidate, Expert$_1$'s position-invariant features enable accurate matching.}
  \label{fig:demo_local2}
\end{figure}

Figure~\ref{fig:demo_global} shows global retrieval results where the system retrieves clinically similar cases based on holistic image features. Figures~\ref{fig:demo_local1} and~\ref{fig:demo_local2} demonstrate local retrieval with bounding-box queries, where HMAR identifies matching lesion regions in candidate images.

\section{Conclusion}
\label{sec:conclusion}

We presented HMAR, an adaptive medical image retrieval framework that unifies global and local retrieval through a Mixture-of-Experts architecture. The dual-expert design---Expert$_0$ for global features and Expert$_1$ for position-invariant local representations---enables multi-granularity retrieval within a single model. A two-stage contrastive training strategy effectively specializes the local expert without requiring bounding-box annotations, while KAN-based hash encoding improves the expressiveness of binary codes. Experiments on the RadioImageNet-CT dataset demonstrate consistent improvements over the state-of-the-art ACIR method.

\paragraph{Limitations and future work.}
The current evaluation is limited to CT images from a single dataset. Future work includes: (1)~extending to additional modalities (MRI, X-ray, ultrasound); (2)~exploring architectures with more than two experts for finer-grained specialization; (3)~developing adaptive gating mechanisms that learn to route based on query content; (4)~incorporating multimodal information (e.g., radiology reports) for richer feature learning; and (5)~conducting large-scale clinical validation studies.

\bibliographystyle{unsrtnat}
\bibliography{refs}

\end{document}